\documentclass[conference]{IEEEtran}

\usepackage{amsfonts}
\usepackage{setspace}

\usepackage{setspace}
\usepackage{amssymb}
\usepackage{epsfig}
\usepackage{graphicx}
\usepackage{xcolor}
\usepackage{algorithm}
\usepackage{algorithmicx}
\usepackage{algpseudocode}
\usepackage{amsmath}

\floatname{algorithm}{Algorithm}

\usepackage[ps2pdf,
bookmarks=false,
bookmarksnumbered=false, 
bookmarksopen=false, 
colorlinks=false]{}

\begin{document}

\title{Adversarial Reinforcement Learning in Dynamic Channel Access and Power Control}

\author{Feng Wang,  M. Cenk Gursoy, and Senem Velipasalar
\\Department of Electrical
		Engineering and Computer Science, Syracuse University, Syracuse, NY, 13244
\\E-mail: fwang26@syr.edu, mcgursoy@syr.edu, svelipas@syr.edu.}

\maketitle

\begin{abstract}
	Deep reinforcement learning (DRL) has recently been used to perform efficient resource allocation in wireless communications. In this paper, the vulnerabilities of such DRL agents to adversarial attacks is studied. In particular, we consider multiple DRL agents that perform both dynamic channel access and power control in wireless interference channels.
	For these victim DRL agents, we design a jammer, which is also a DRL agent. We propose an adversarial jamming attack scheme that utilizes a listening phase and significantly degrades the users' sum rate. Subsequently, we develop an ensemble policy defense strategy against such a jamming attacker by reloading models (saved during retraining) that have minimum transition correlation.
\end{abstract}

\begin{IEEEkeywords}
	Adversarial policies, defense strategies, dynamic channel access, deep reinforcement learning, jamming attacks.
\end{IEEEkeywords}

\section{Introduction}
The limited nature of spectral and energy resources in wireless systems along with time-varying channel conditions necessitate dynamic operation with spectral and energy efficient transmission schemes \cite{chen2012stochastic}. As a data driven approach, deep reinforcement learning (DRL) addresses complex control problems by adapting to dynamic environments and making sequential decisions. DRL has recently been applied in several domains, including game playing \cite{silver2017mastering}, autonomous driving \cite{sallab2017deep}, robotics \cite{gu2017deep}, computer vision \cite{yun2017action}, etc. With these, there is growing interest in employing DRL in communication systems to address, e.g., modulation classification \cite{8645696}, dynamic channel access \cite{8303773}, and power allocation \cite{nasir2019multi}. In particular, interference management is one of the most demanding challenges in wireless communications and channel  allocation and transmission power control has been extensively studied in the literature (see e.g., \cite{chiang2008power} \cite{kiani2007maximizing} and \cite{zhang2011weighted}). As noted above, recently DRL has been introduced to address such resource allocation problems. For instance, the authors in \cite{nasir2019multi} proposed a distributively executed deep Q-learning algorithm for transmitters to decide their transmission power, without any knowledge of channel state information or a given training data set generated by a near-optimal algorithm.

Recently, in the presence of maliciously crafted adversaries, machine learning is shown to be vulnerable against manipulated feedback \cite{goodfellow2014explaining}, e.g., in image classification \cite{dong2018boosting,dong2019evading,jia2019enhancing}, game playing \cite{lin2017tactics}, time series classification \cite{fawaz2019adversarial}, and in reinforcement learning models as well \cite{huang2017adversarial,xiao2019characterizing,zhao2019blackbox}. In aforementioned studies, the victim models are massively trained and the parameters are fixed after training. The inputs and outputs of victim models are also open to the adversary. However, in a wireless setting, an adversarial jamming attacker have limited observation on the environment and limited knowledge of the feedback from the dynamic users. Furthermore, the victim users' decision-making DRL model is computationally adept enough to retrain and adapt to the changing environment, which makes it infeasible for an outer jamming attacker to access the victims' DRL model parameters to perform any gradient-based attack. To overcome this limitation, in \cite{gleave2019adversarial}, the authors trained another adversarial DRL agent whose reward is a function of victim's reward. Based on this, jamming attack on a link with one pair of transmitter and receiver in a single channel is studied in \cite{sagduyu2019adversarial} and \cite{erpek2018deep}. In \cite{zhong2020adversarial}, we considered a scenario with multiple channels, and studied the interaction between jamming attacker and victim user, performing DRL-based dynamic channel access. We analyzed the attack strategy as a dynamic control problem, and proposed a DRL jamming attacker which utilizes a periodical listening phase to perform effective attacks.

Motivated by these adversarial attack strategies, there has been increasing focus on adversarial training to improve deep learning agents' robustness against minor but effective adversarial perturbation \cite{kurakin2016adversarial,madry2017towards,zhang2019theoretically}. However, DRL could be more fragile. Even without attack, DRL agents can make catastrophic mistakes on extremely unexpected inputs \cite{uesato2018rigorous}. Therefore, robust deep reinforcement learning strategies have been proposed to improve DRL performance in the worst case \cite{zhang2020robust}. These aforementioned works, addressing whether white-box or black-box attacks, assume that the attacker's model or its feedback is fully accessible. However for jamming attacks and anti-jamming defensive mechanisms, both victim and attacker have limited observations of each other. Defense strategies against jamming attack are thus studied \cite{sagduyu2020wireless}. While the authors in \cite{erpek2018deep} and \cite{shi2018adversarial} investigated mitigating wireless jamming attacks in one channel, scenarios with multiple channels are studied in \cite{wang2020defense} and \cite{wang2020adversarial}, where an ensemble of several orthogonal policies is generated and used as a defense strategy against jamming attacks.

In this paper, we consider multiple dynamic channels \cite{nasir2019multi} and a joint power allocation and dynamic channel access problem in a network with multiple links. The victim users' deep Q-network (DQN) for decision making is well-trained to achieve high sum-rate \cite{lu2019dynamic}. We first propose a DQN-based jamming attacker that has the same power limit as the victim users, and dynamically adapt to the environment and victims' policies to minimize the victim users' sum-rate. Subsequently, we design a defense strategy to maximize the victim users' sum-rate. We show that simply retraining the victims' DQN agent may fail catastrophically, and we develop a defensive scheme to reload the saved models with minimum correlation in the transitions between states.

\section{Dynamic Channel Access and Power Control Policies of the Victim User}\label{Sec: pre}
In this section, we introduce the background on DQN based dynamic multichannel access and power control. As noted above, we consider a dynamic environment proposed in \cite{nasir2019multi} and the DQN agent proposed in \cite{lu2019dynamic} as the victim user.

\subsection{Dynamic Interference Channel Model}\label{subsec: channel}
We consider a wireless network model with $K$ transmitter-receiver pairs. There are $N_c$ channels. Let $h_{ij}^{(c)} \sim \mathcal{CN} (0,1)$ denote the fading coefficient of the link from transmitter $j$ to receiver $i$ in channel $c$, which is circularly symmetric complex Gaussian (CSCG) distributed with zero mean and unit variance. The fading coefficients vary every $T$ time slots according to the Jakes fading model \cite{liang2017spectrum}, and we express them as a first-order complex Gauss-Markov process as in \cite{nasir2019multi}:
\begin{equation} \label{eq:fading}
h_{ij}^{(c)}((n+1)T) = \rho h_{ij}^{(c)}(nT) + \sqrt{1-\rho^2} e_{ij}^{(c)}((n+1)T)
\end{equation}
where $n$ is a positive integer, $h_{ij}^{(c)}(t)$ is the fading coefficient at time $t$, the channel innovation process $e_{ij}^{(c)}(t)$ is independent and identically distributed CSCG random variable, and $\rho=J_0(2\pi f_d T)$, where $J_0$ is zeroth order Bessel function of the first kind and $f_d$ is the maximum Doppler frequency.

For transmission, each user chooses one channel from $N_c$ channels. The transmit power has $N_p$ levels, and the power of the $k$th transmitter is denoted as $P_k \in \{\frac{P_{\max}}{N_p},\frac{2P_{\max}}{N_p},\ldots,P_{\max}\}$. Therefore, the received signal-to-interference-plus-noise-ratio (SINR) for the $k$th transmitter-receiver pair that has chosen channel $c$ can be expressed as
\begin{equation} \label{eq:SINR}
\text{SINR}_k = \frac{P_k |h_{kk}^{(c)}|^2}{\sum_{j\neq k}{P_j |h_{kj}^{(c)}|^2}+\sigma_k^2}
\end{equation}
where $\{j\neq k\}$ denotes the indices of the transmitters that have selected channel $c$, $\sigma_k^2$ is the variance of the additive white Gaussian noise at receiver $k$. Now, we express the individual transmission rate $r_k$ of the $k$th user pair and the sum rate $r$ as follows:
\begin{equation} \label{eq:indirate}
r_k = \log_2(1+\text{SINR}_k),
\end{equation}
\begin{equation} \label{eq:sumrate}
r = \sum_{k=1}^{K}r_k.
\end{equation}

Denoting the selected channel and power level of user pair $k$ by $c_k$ and $P_k$, respectively, we can express the sum-rate maximization problem as
\begin{equation} \label{eq:sumrateopt}
\operatorname*{argmax}_{P_1,\ldots,P_K,c_1,\ldots,c_K}{\sum_{k=1}^{K} {\log_2\left(1+\frac{P_k |h_{kk}^{(c_k)}|^2}{ \sum_{j\neq k}{\mathbf{1}_{jk} P_j|h_{kj}^{(c_k)}|^2}+\sigma_k^2}\right)}}
\end{equation}
where $\mathbf{1}_{jk}$ is the indicator function $\mathbf{1}(c_j=c_k)$.

\subsection{DQN Agent}\label{subsec: victim}
To solve the challenging problem described in (\ref{eq:sumrateopt}), Lu et al. in \cite{lu2019dynamic} constructed multi-agent deep reinforcement learning algorithms to jointly perform dynamic channel access and power control in interference channels to decide channel and power selection for each user pair with the goal to maximize the sum-rate. It is assumed that the fading coefficients $h_{ij}^{(c)}$ are unknown to the users, and the feedback from environment includes the individual transmission rates. Such dynamic channel access and power control with limited knowledge can be modeled as a partially observable Markov decision process (POMDP).

The users observe the state at the end of each time slot, and we denote the observation at time $t$ as $O(t) = \{o_1(t), o_2(t), \ldots, o_K(t) \}$. For user pair $k$, $o_k(t)$ includes one entry indicating its individual rate, and $N_p$ entries denoting transmission power $P_k$ in each channel. The transmission power ranges from $0$ to $P_{\max}$. Note that we have modified the representation of state observation as in \cite{lu2019dynamic} with smaller number of DQN input entries to carry the same information, in order to accelerate the DQN training process and enhance the performance. If we keep $N$ observations from the past, then the observation space at time $t$ can be denoted as $\mathcal{O}(t) = \{O(t-1), O(t-2), \ldots, O(t-N) \}$.

We consider the discrete action space $\mathcal{A} = [\mathcal{C}, \mathcal{P}]\bigcup\{\text{no transmission}\}$ for each user pair at each time, where $\mathcal{C}\in\{1, \ldots, N_c\}$ is the set of actions for channel selection and $\mathcal{P}\in\{1, \ldots, N_p\}$ is the set of actions for power levels. Hence, there are $N_c \times N_p$ actions through which the transmitter selects the channel and transmission power level, and there is one action that indicates no transmission. Therefore, there are overall $N_c \times N_p + 1$ possible actions. 

The reward $r(t)$ is the sum rate in (\ref{eq:sumrate}) at time $t$. The aim is to find an optimal policy $\pi^*$ among all feasible policies. For each user pair, the policy maps the observation space $\mathcal{O}(t)$ to action space $\mathcal{A}$ to jointly maximize the discounted sum reward:
\begin{equation} \label{eq:policyopt}
\pi^* = \operatorname*{argmax}_{\pi}{\sum_{t=0}^{T'}{\gamma ^t r_{t+1}}}
\end{equation}
where $r_{t}$ denotes the sum-rate at time $t$ and $\gamma$ is the discount factor.

To estimate the expected reward for each possible action, we employ a deep neural network to estimate the Q-value. The input states first go through a long short term memory (LSTM) layer \cite{hausknecht2015deep} to estimate the underlying state of this POMDP, and then pass a dueling network to avoid states that potentially lead to poor performance \cite{wang2016dueling}. We assume that the central unit trains the DQN with information from all users, and then sends its well-trained copies to each transmitter. Subsequently, the transmitters can make decisions in a distributed fashion with their own local information.

\subsection{Performance Evaluation in the Absence of Jamming Attacks}\label{subsec: vicexp}

We consider the dynamic fading interference channel described in Section \ref{subsec: channel}, and evaluate the sum rate achieved with the DQN agents in Section \ref{subsec: victim}. First, the central unit (CU) trains the DQN with feedback from all user links. During $\epsilon$-greedy training, each user link has probability $\epsilon$ to explore a random action, and has probability $(1-\epsilon)$ to choose the action with the maximum Q-value from DQN. The value of $\epsilon$ decreases linearly from $\epsilon_0$ to $\epsilon_1$ over the training time $T_{train}$. When the training time is over, CU distributes a copy of DQN to each user, and they start making decisions in a distributed fashion and do not train the DQN parameters any further.

\begin{figure}
	\centering
	\includegraphics[width=0.7\linewidth]{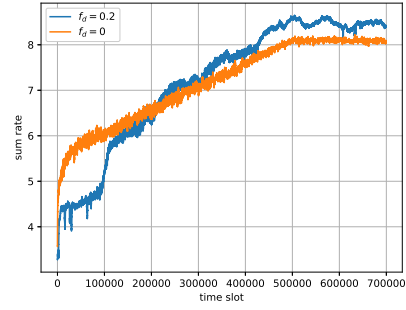}
	\caption{Sum rate of the victim user in the absence of jamming attacks.}
	\label{fig:NoAttack_curve}
\end{figure}

\begin{figure}
	\centering
	\includegraphics[width=0.7\linewidth]{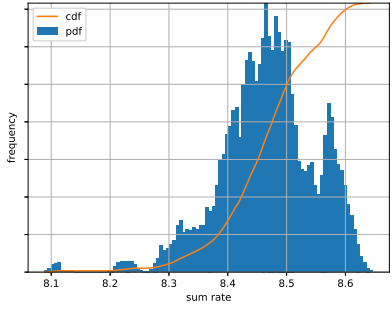}
	\caption{Empirical PDF and CDF of sum rate in a dynamic environment with $f_d = 0.2$ after training is completed at $t=500000$.}
	\label{fig:NoAttack_CDF}
\end{figure}

In our simulations, we assume there are $K=2$ transmitter-receiver pairs and $N_c=4$ channels to better evaluate our attack and defense strategies. All other parameters are shown in Table \ref{tab:vic}. This DQN model is evaluated in the absence of jamming attacks. In Fig. \ref{fig:NoAttack_curve}, we compare the moving average sum rates of the model trained in a dynamic environment with $f_d=0.2$ and those of the model in a fixed environment with $f_d=0$. We see that the model in the dynamic environment is forced to explore and react to different channel states, has a better chance to attain an improved performance level, and results in a higher sum rate. The empirical probability density function (PDF) and empirical cumulative distribution function (CDF) of the sum rate during test time after $t>5 \times 10^5$ are plotted in Fig. \ref{fig:NoAttack_CDF}. We see that the sum rate concentrates around 8.5. For the simulations in the following sections, we consider a dynamic environment with $f_d=0.2$.

\begin{table}[h!]
\small
	\begin{center}
		\caption{Victim User Training Parameters}
		\label{tab:vic}
		\begin{tabular}{ | l | l |}
			\hline
			Doppler Frequency $f_d$ & 0.2 \\ \hline
			Dynamic Slot Duration $T$ & 0.02 \\ \hline
			Gaussian Noise Power $\sigma_k^2$ & 1   \\ \hline
			Number of User Links $K$ & 2  \\ \hline
			Number of Channels $N_c$ & 4 \\ \hline
			Maximum Power $P_{\max}$ & 38dBm (6.3W) \\ \hline
			Number of Power Levels $N_p$ & 5 \\ \hline
			Discount Factor $\gamma$ & 0.9 \\ \hline
			Learning Rate & 0.04 \\ \hline
			LSTM Hidden Layer Nodes & 20 \\ \hline
			Duel Network Hidden Layer Nodes & 10 \\ \hline
			Initial Exploration Probability $\epsilon_0$ & 1 \\ \hline
			Final Exploration Probability $\epsilon_1$ & 0.1 \\ \hline
			Training Time $T_{train}$ & 500000 \\ \hline
			Testing Time $T_{test}$ & 200000 \\ \hline
		\end{tabular}
	\end{center}
\end{table}


\section{Jamming Attacker}\label{sec: attacker}
In this section, we design a DQN agent to perform jamming attacks on the aforementioned victim users. We assume that the attacker has no prior information on the channel environment or the victims' DQN parameters, such as the weight and biases. The attacker is able to jam $K_a$ channels simultaneously if it chooses to attack, or observes the interference plus noise with its receiver if it chooses to listen. We also assume that the attacker can eavesdrop on the victim user links' to learn the sum rates from the receivers' feedback at the end of each time slot. We demonstrate that a DQN attacker under such assumptions can significantly degrade the sum rate of the users that employ the DQN agents described in the previous section.

\subsection{DQN agent}\label{subsec: attagent}
If we describe the attacker as an additional user that can choose $K_a$ channels to attack simultaneously but does not contribute to the sum rate, we can formulate the attacker's sum-rate minimization problem by rewriting (\ref{eq:sumrateopt}) as
\begin{equation} \label{eq:SINRatt}
\begin{aligned}
\operatorname*{argmin}_{P_{(K+1)1},\ldots,P_{(K+1)K_a},c_{(K+1)1},\ldots,c_{(K+1)K_a}}\sum_{k=1}^{K} \log_2 \\
\left(1+\frac{P_k |h_{kk}^{(c_k)}|^2}{\scriptstyle \sum_{j=1}^{K,j\neq k}{\mathbf{1}_{jk} P_j|h_{kj}^{(c_k)}|^2}+\sigma_k^2+\sum_{i=1}^{K_a}{\mathbf{1}_{ik}P_{(K+1)i} |h_{k(K+1)}^{(c_k)}|^2}}\right)
\end{aligned}
\end{equation}
where $c_{(K+1)i}$ denotes the $i$th channel that the attacker jams, $P_{(K+1)i}$ denotes the jamming power in channel $c_{(K+1)i}$, $h_{k(K+1)}^{(c_k)} \sim \mathcal{CN} (0,1)$ denotes the fading coefficient from the attacker's transmitter $(K+1)$ to victim user's receiver $k$ in channel $c_k$. $\mathbf{1}_{ik}$ is the indicator function $\mathbf{1}(c_i=c_{(K+1)k})$.

Since the attacker has a similar impact on the environment and has a similar objective function, we deploy another DQN agent which has the same structure as the victim DQN agent without acquiring any parameters. The attacker and the victim user choose the channels to transmit or listen at the same time, and then the attacker collects its reward at the end of the time slot for training the model and optimizing its policy. The attacker can be in two different operational modes: attacking mode and listening mode.

\subsubsection{Attacking mode}\label{subsec: attmode}
In this mode, the attacker selects and jams $K_a$ channels, and eavesdrops to learn the victims' sum rate to update its neural network. The attacker agent always attacks $K_a$ different channels with full power $P_{\max}$. Therefore, attacker agent's action space $\mathcal{A}_a$ is the combination of $K_a$ channels out of all $K$ channels, i.e., there are ${K}\choose{K_a}$ different actions. The only accessible information in attacking mode is the victims' sum rate $r$ in (\ref{eq:sumrate}). Based on this, we define the observation, which is the input of the attacker DQN, as $\mathcal{O}_a(t) = \{O_a(t-1), O_a(t-2), \ldots, O_a(t-N) \}$. The observation $O_a$ at a given time consists of one entry indicating the victims' sum rate $r$, and $N_c$ entries, each of which indicates whether the corresponding channel was jammed or not with values 1 or 0, respectively. The reward of the attacker's DQN agent is the negative of the sum rate, $-r$. The attacker's policy is updated with the observation, action and reward to adapt to the victim and the dynamic environment. Although the attacker jams the chosen channels in the attacking mode and degrades the victims' performance, it collects only limited information regarding the victims' choices due to its inability to listen during jamming attacks. Therefore, the attacker sometimes switches to the listening mode.

\subsubsection{Listening mode}\label{subsec: lismode}
In this mode, the attacker does not jam any channel, and use its receiver to measure the interference plus noise. We assume that the noise has a similar distribution over all channels, and therefore the channels with higher interference plus noise have a higher probability to have been selected by victims for transmission. We record $K_a$ channels with the highest interference levels as the chosen action, and set $0$ as the reward which is higher than $-r$ in attacking mode. Therefore, we can further update the attacker's policy to choose these observed victim actions, when the same observation history occurs again. However, the attacker does not jam any channel to affect the victim, and we record $K_a$ zeros in observation history $O_a(t)$ for the following policy training.

On the one hand, to suppress the victims' sum rate, the attacker tends to choose the same channels as the victims do. If a listening mode history appears in the observation history $\mathcal{O}_a$ of another time slot, such observation is less practical for training the policy. Therefore, the attacker should not use listening mode successively. On the other hand, it is rare for the attacker to coincide with the victim in attack mode at the beginning of training, while listening mode will guarantee an accurate observation on victims' chosen channels. To strike a balance between the drawbacks and benefits of listening, we propose an $\epsilon_l$-greedy operation mode. In this mode, the attacker chooses the listening mode with probability $\epsilon_l$ that decreases over the training time, and switches to attack mode if it has chosen the listening mode in the last time slot. If it chooses the attack mode, it still has an $\epsilon$-greedy method to explore random actions. We show this $\epsilon_l-\epsilon$ greedy exploration method in Algorithm \ref{alg:opmode} below.

\begin{algorithm}
\small	
\caption{$\epsilon_l-\epsilon$ greedy exploration method}
	\label{alg:opmode}
	\begin{algorithmic}
		\State {$\epsilon_l=\epsilon_{l0}$}
		\State {$\epsilon=\epsilon_{0}$}
		\For{i in range($T_{train}$)}
			\If {random$(0,1)<=\epsilon_l$}
				\State{Victim execute listening mode}
			\Else
				\If {random$(0,1)<=\epsilon$}
					\State{Victim execute attacking mode}
				\Else
					\State{Victim explore random action}
				\EndIf
			\EndIf
			\State {$\epsilon_l=\epsilon_l-(\epsilon_{l0}-\epsilon_{l1})/T_{train}$}
			\State {$\epsilon=\epsilon-(\epsilon_{0}-\epsilon_{1})/T_{train}$}
		\EndFor
	\end{algorithmic}
\end{algorithm}

We should also notice the consequences of operating in a dynamic environment.  When reward is high, the victim user just stays in the same state. If reward drops, it notices that the environment has changed, and switches to other states. If one user fails to choose the channel that supports the highest transmission rate, its individual rate will not decrease significantly if the actually chosen channel is idle. However, for the attacker, its optimal state is affected by both the environment and victim. If it once fails to choose exactly the same channels as the victims do, the victims' sum rate increases sharply. Therefore, the attacker should always train its model to adapt to the environment and the victim, even after the $\epsilon_l-\epsilon$ greedy method is over.

\subsection{Simulations}\label{subsec: attexp}
In this section, we test the proposed DQN jamming attacker against the victim agents in the wireless interference channel introduced in Section \ref{Sec: pre}. We assume the victims' policy is well-trained before the attacker begins to observe and attack. In our simulations, we assume that the Doppler frequency is $f_d=0.2$, and the attacker can jam $K_a=2$ channels simultaneously. When the attacker trains its neural network, it bootstraps 16 pairs of observation-action-reward from the history of 10000 time slots. All other parameters are listed below in Table \ref{tab:att}.

\begin{table}[h!]
\small
	\begin{center}
		\caption{Attacker Training Parameters}
		\label{tab:att}
		\begin{tabular}{ | l | l |}
			\hline
			Gaussian Noise Power $\sigma_k^2$ & 1   \\ \hline
			Number of Attacker Link & 1  \\ \hline
			Number of Attacked Channels $K_a$ & 2  \\ \hline
			Jamming Power $P_{\max}$ & 38dBm (6.3W) \\ \hline
			Discount Factor $\gamma$ & 0.9 \\ \hline
			Learning Rate & 0.2 \\ \hline
			LSTM Hidden Layer Nodes & 20 \\ \hline
			Duel Network Hidden Layer Nodes & 10 \\ \hline
			Initial Exploration Probability $\epsilon_0$ & 1 \\ \hline
			Final Exploration Probability $\epsilon_1$ & 0.1 \\ \hline
			Initial Listen Probability $\epsilon_{l0}$ & 0.25 \\ \hline
			Final Listen Probability $\epsilon_{l1}$ & 0.025 \\ \hline
			Initial Training Time $T_{train}$ & 20000 \\ \hline
		\end{tabular}
	\end{center}
\end{table}

\begin{figure}
	\centering
	\includegraphics[width=0.7\linewidth]{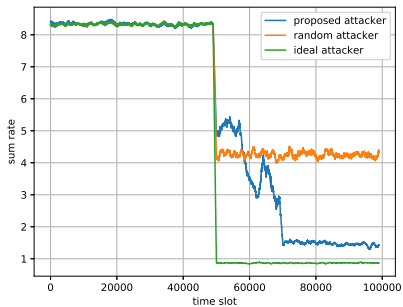}
	\caption{Sum rate of the victim user under jamming attacks.}
	\label{fig:Attackers_curve}
\end{figure}

\begin{figure}
	\centering
	\includegraphics[width=0.7\linewidth]{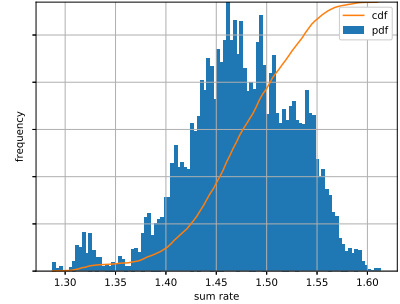}
	\caption{Empirical PDF and CDF of sum rate under proposed attacker after training is done at $t=70000$.}
	\label{fig:Attackers_CDF}
\end{figure}

In Fig. \ref{fig:Attackers_curve}, in terms of victims' sum rate, we compare the proposed attacker against a random attack which randomly selects $K_a=2$ channels to jam each time, and an ideal attacker which always selects the same channels as the victims do. The attackers do not observe or attack until $t=50000$, and then they begin to attack, and the victims' sum rate drops significantly. The proposed attacker executes the $\epsilon_l-\epsilon$ greedy method during $50000\le t<70000$, and then the probabilities are fixed at $\epsilon_1$ and $\epsilon_{l1}$, respectively. The ideal attacker is infeasible to apply in practical scenarios, and we provide its performance as an upper bound on the performance of any possible attacker policy. Even though the proposed attacker has no prior knowledge on the victim agent and ceases to attack in the listening phase, we observe that the converged sum rate in the presence of the proposed attacker is very close to that with the ideal attacker, while both are much lower than what is achieved with the random attacker. We notice that the proposed attacker performs worse than the random attacker at the beginning of the training, because it has high exploration probability $\epsilon$ with which it chooses randomly, and listens with probability $\epsilon_{l1}$ during which it does not attack. We should also notice that the training time of the proposed attacker is only 20000 time slots, which is less than $5\%$ of the victims' training duration of 500000 time slots. The PDF and CDF of victims' sum rate under the proposed attack after $t=70000$ is shown in Fig. \ref{fig:Attackers_CDF}. We observe that the converged sum rate is less than 1.5, which is substantially less than the sum rate of 8.4 before attack.


\section{Ensemble Policy Defense}\label{sec: defense}
In the previous section, we show that a jamming attack with the same power can significantly reduce the victims' sum rate. Different from the previous work on ensemble policy \cite{hans2010ensembles} and pruning in terms of performance \cite{partalas2006ensemble}, we in \cite{wang2020adversarial} provided a defense strategy that utilizes two orthogonal policies. The goal has been to maximize the difference between policies in the ensemble in order to achieve good performance even under attack. However in the considered interference channel, there is no guarantee that multiple orthogonal policies will find idle channels with high SINR. Therefore, we here introduce an ensemble of policies with minimum transition correlation. We note that this proposed ensemble policy does not make any assumption on the attacker policy, and learns only from the interactions. Therefore, it could be an universal solution to different types of attack.

One intuitive approach to mitigate the jamming attack is that the CU trains the victims' DQN again. However, the scenario in which $K=2$ victim user pairs are attacked by a jammer that attacks $K_a=2$ channels, is not the same as the scenario where 4 user pairs work cooperatively. In the latter case, all user agents explore different channels to maximize the sum rate. However, in the former scenario, attacker intends to maliciously attack the channels chosen by the victims. There are only $N_c-K_a=2$ channels that are not jammed, and conditions of the channels vary over time as the attacker adapts to the victims' choices. We should also notice that these unattacked channels do not guarantee high transmission rate because of the dynamic channel conditions and the potential interference when two or more users select the same channel for transmission. Therefore, the action that results in only one user transmitting may often receive high reward, and the DQN may also accumulate high reward with the action of "no transmission". Finally, the victim DQN model collapses because it chooses "no transmission" much more often than transmission in a channel, resulting in very low sum rates.

\begin{figure}
	\centering
	\includegraphics[width=0.8\linewidth]{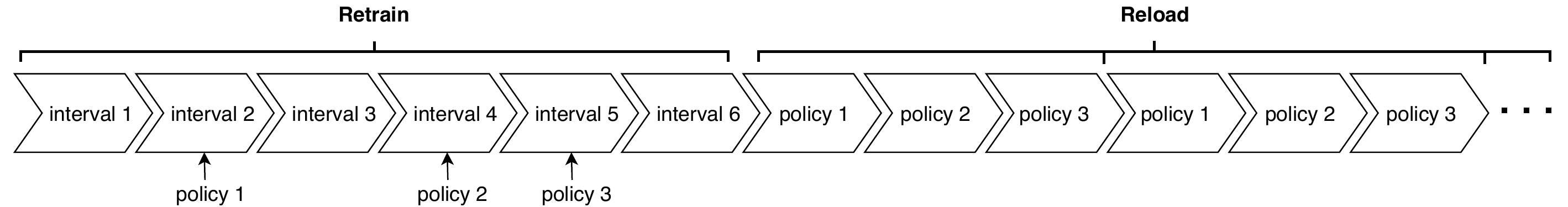}
	\caption{Model collection and usage for ensemble policy.}
	\label{fig:EnsemblePolicy}
\end{figure}

To make use of the recorded interactions between the victim agents and the attacker agent, we propose a defense strategy that utilizes the saved policies with minimum correlation value between transition matrices as depicted in Fig. \ref{fig:EnsemblePolicy}. We assume that the CU retrains the DQN model while being attacked, saves $N_s$ different DQN models with fixed time interval $T_e=T_{retrain}/N_s$ over a total duration of $T_{retrain}$, and then stops before it collapses. At the $t$th time slot of the $n_T$th time interval, let $a_1$, $a_2$ indicate the channels selected by the victim at times $t-1$ and $t$ respectively, and let $r_1$ be the quantized integer value indicating victim reward at time $t-1$. Now, we define each element in the first order transition matrix $M_{n_T}(a_1,a_2,r_1)$ as
\begin{equation} \label{eq:transmat}
\begin{aligned}
M_{n_T}(a_1,a_2,r_1) = \sum_{k=1}^{K_a}\sum_{t=2}^{T_e}{\Big(\mathbf{1}(c_k(t-1)=a_1)}\\
	{\mathbf{1}(c_k(t)=a_2)\mathbf{1}(r_1<=r(t-1)<r_1+1)\Big)}
\end{aligned}
\end{equation}
where $c_k(t)$ is the channel chosen by transmitter $k$ at time $t$, and $r(t)$ is the victim reward at time $t$. We then define the transition correlation between the first order transition matrices $M_{n_1}$, $M_{n_2}$ as
\begin{equation} \label{eq:corr}
R(n_1, n_2) = M_{n_1} \circ M_{n_2}
\end{equation}
where $\circ$ denotes element-wise Hadamard product.

For the $n_T$th time interval and the corresponding DQN model saved during that time interval, we have the transition correlation curve between $M_{n_T}$ and $M_{n}, n=1,2,\ldots,N_s$. If we put all curves for all $N_s$ time intervals together, we can distinguish $N_e$ different intervals that have relatively lower correlation to others. These $N_e$ corresponding DQN models may lead to maximally different transition, and it is more difficult for the attacker agent to adapt to this subset of policies.

To take advantage of this, CU will distribute these $N_e$ DQN models to all users, and they will in turn make use of the ensemble of policies. Starting from the time that the models are distributed, users make decisions and retrain models with local information. Additionally, they reload the parameters of the next model in sequence with a fixed time interval periodically, so that they go over all $N_e$ models with period $T_{reload}$. Therefore, each model only experiences a short time of retraining, and is unlikely to collapse. Another benefit is that the system still runs in a distributed fashion after $N_e$ models are collected at the CU.

To effectively use the ensemble of policies, we should have $T_{reload} < T_{retrain}$. Therefore, we accelerate the rotation in transition space between policies, which makes it harder for the attacker to imitate such an ensemble of policies than imitating the retraining at the CU.

\section{Simulations with Ensemble Policy Defense}\label{sec: exp}
In this section, we test the ensemble policy approach as a defense against the attacker proposed in Section \ref{sec: attacker}. In all simulations in this section, the victim is well-trained at the beginning, and the attacker starts attacking at $t=50000$, and its $\epsilon_l-\epsilon$ greedy method is completed at $t=70000$. We assume that the victim notices the attack and starts the centralized training at $t=100000$.

\begin{figure}
	\centering
	\includegraphics[width=0.7\linewidth]{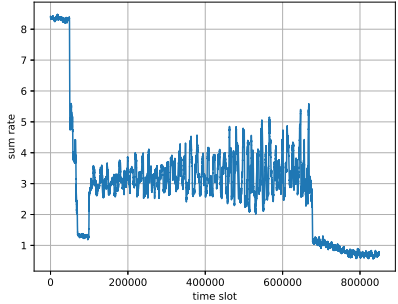}
	\caption{Sum rate of the victim user under attack fails during retraining.}
	\label{fig:Victimfail}
\end{figure}

Although we cannot terminate the retraining of the victim policy because of the adaptive attacker, we observe that the victim model might collapse after a long period of retraining. As we see in Fig. \ref{fig:Victimfail}, after the victim starts to retrain at $t=100000$, its policy and the sum rate begin to oscillate, until it collapses after $t=600000$. Due to the random initialization of the neural network parameters, the time of collapse can be different among models with different initial parameters.

\begin{figure}
	\centering
	\includegraphics[width=0.7\linewidth]{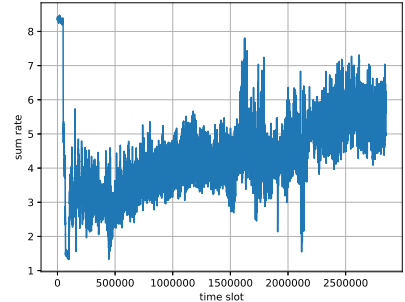}
	\caption{Sum rate of the victim user under attack with ensemble policy.}
	\label{fig:ensemble}
\end{figure}

In the previous section, we have described the ensemble policy with minimum transition correlation. Therefore, we no longer perform long-term training after we obtain $N_e$ policies. In Fig. \ref{fig:ensemble}, we show the performance in terms of victim sum rate when the ensemble policy defense is employed. In the time interval $100000\le t<1550000$, the victim CU retrains the model, and the model begins to collapse at $t=1550000$. At this time, the CU distributes $N_e=8$ models to all users, each of which trains and reloads periodically with reloading period $T_{reload} = 720000$. All other parameters are shown below in Table \ref{tab:def}.

\begin{table}[h!]
\small	
\begin{center}
		\caption{Ensemble Policy Parameters}
		\label{tab:def}
		\begin{tabular}{ | l | l |}
			\hline
			Learning Rate & 0.4 \\ \hline
			Exploration Probability $\epsilon$ & 0.05 \\ \hline
			Retraining Time $T_{retrain}$ & 1450000 \\ \hline
			Number of intervals $N_s$ & 72 \\ \hline
			Number of Ensembled Policies $N_e$ & 8 \\ \hline
			Reloading Period $T_{reload}$ & 720000 \\ \hline
		\end{tabular}
	\end{center}
\end{table}

\begin{figure}
	\centering
	\includegraphics[width=0.7\linewidth]{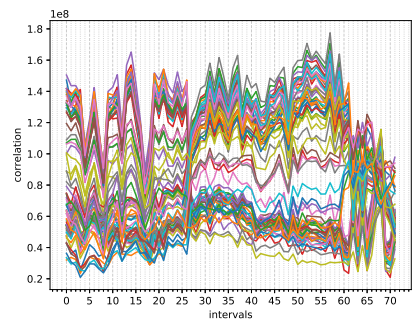}
	\caption{1-dimensional plot of transition correlation matrix during retraining.}
	\label{fig:transmat}
\end{figure}

\begin{figure}
	\centering
	\includegraphics[width=0.7\linewidth]{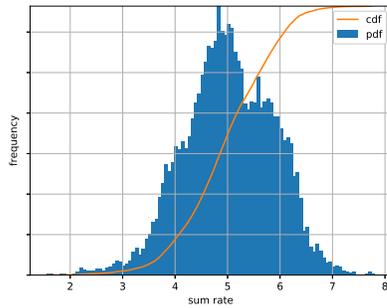}
	\caption{CDF and PDF of victim sum rate with ensemble policy.}
	\label{fig:ensembleCDF}
\end{figure}

Based on the central training history during $100000\le t<1550000$, we extract the row vectors of the transition correlation matrix, and plot them as line curves to represent the trends in Fig. \ref{fig:transmat}. We can see the model collapsing in Fig. \ref{fig:ensemble} at $t=1550000$ as well as in Fig. \ref{fig:transmat} in the last few intervals. Therefore, we only collect $N_e=8$ models with minimum correlation from the first 65 intervals. Then in Fig. \ref{fig:ensemble}, we start to reload these models with period $T_{reload} = 720000$. We can see an upsurge of sum rate at about $t=2400000$, which means that the attacker can hardly learn from the interactions with the victim after the second reloading period starts. We plot the CDF and PDF of the performance achieved with the ensemble policy defense after $t=1550000$ in Fig. \ref{fig:ensembleCDF}. We see that the average of sum rate exceeds 5, which is even higher than the victim under random attack as shown in Fig. \ref{fig:Attackers_curve}. The jamming attacker achieving a lower sum rate than that of the random attacker indicates the ineffectiveness of attacker's DQN agent once the victims employ the ensemble policy. Therefore, the proposed ensemble policy defense successfully mitigates such jamming attacks.

\begin{figure}
	\centering
	\includegraphics[width=0.7\linewidth]{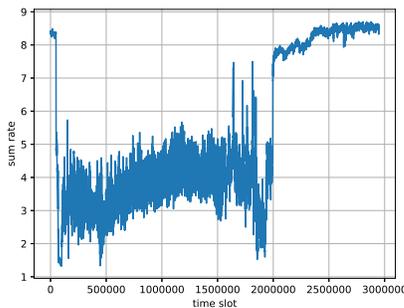}
	\caption{Sum rate of the victim user under attack with ensemble policy. Due to attacker's initialization, its DQN agent converges to a catastrophic local maximum, and barely selects the same channels as victim does.}
	\label{fig:Victim_success}
\end{figure}

Additionally, we find a special case during our simulations that the attacker DQN's random initialization is poor, and converges to a catastrophic local maximum. This only happens when the attacker encounters victims' ensemble policy, and we show this result in Fig. \ref{fig:Victim_success}. All other parameters are the same as in Fig. \ref{fig:ensemble}, and we see the victim sum rate surges because the attacker collapses at $t=2000000$.

\section{Conclusion}\label{sec: con}
In this paper, we have introduced a DQN jamming attacker that has the goal to minimize the victims' sum rate while being subject to the same power constraint as the victim users. We have explicitly designed its architecture and described its operation modes, and demonstrated its successful performance against the victim agents in absence of defense strategies. We have then provided explanations on why the victim should not retrain indefinitely under attack, and proposed an ensemble policy with minimum transition correlation as a defense strategy. We have defined the transition matrix and the correlation, and investigated how to extract a subset of policies with minimum correlation in practice. We have shown the effectiveness and success of the proposed defensive strategy against the proposed DQN attacker by quantifying the improvements in the sum rates and providing comparisons with the sum rate achieved under random attacks.

\bibliographystyle{IEEEtran}
\bibliography{ref_AA}

\end{document}